\pdfoutput=1
\documentclass{article}
\usepackage{spconf,amsmath,graphicx}
\usepackage{comment}
\usepackage{tabularx}
\usepackage{booktabs}
\usepackage{multirow}
\usepackage{multicol}
\usepackage{xcolor}
\usepackage{soul}
\usepackage{booktabs}
\usepackage{multirow}

\title{Generalising deep learning MRI reconstruction across different domains}

\name{\parbox{\linewidth}{\centering 
Cheng Ouyang$^{1}$\thanks{This study is supported by an EPSRC programme grant (EP/P001009/1).} 
Jo Schlemper$^{1}$ 
Carlo Biffi$^{1}$ 
Gavin Seegoolam$^{1}$  
Jose Caballero$^{1}$ \\
Anthony N. Price$^{2}$ 
Joseph V. Hajnal$^{2}$ 
 Daniel Rueckert$^{1}$ }}

\address{$^{1}$Biomedical Image Analysis Group, Department of Computing, Imperial College London, UK\\
$^{2}$Division of Imaging Sciences and Biomedical Engineering Department, King’s College London, UK}

\begin{document}
\ninept
\maketitle

\begin{abstract}
\ninept
We look into robustness of deep learning based MRI reconstruction when tested on unseen contrasts and organs. We then propose to generalise the network by training with large publicly-available natural image datasets with synthesised phase information to achieve high cross-domain reconstruction performance which is competitive with domain-specific training. To explain its generalisation mechanism, we have also analysed patch sets for different training datasets.
\end{abstract}

\section{INTRODUCTION}
\label{sec:intro}
\ninept

A deep learning based reconstruction model trained for a specific scanning setting (\textit{i.e.} a domain) usually underperforms on unseen contrasts or organs due to the domain shift problem. As obtaining fully-sampled images for each domain is impractical, we propose a simple generalisation strategy for deep MRI reconstruction. 

\section{METHODS and experiments}
\label{sec:meth_exp}
\ninept
We build the strategy on one of the state-of-the-art deep-cascade of CNN\cite{schlemper2018deep}, which learns fully-sampled image priors and projects undersampled images to the learned fully-sampled image space. 

We train the network with MS-COCO Stuff Segmentation dataset\cite{lin2014microsoft}, which contains around 118k natural images ($>$100$\times$ larger than any of three MRI datasets used in this work which contain 0.3k to 1k 2D slices each). Synthetic phases are then added, which is crucial for model sharing between natural images and MRI.

For characterisation of domain shift and for evaluation, we have performed comparisons across different domains (training and testing on different domains) for 2D single-coil slice-by-slice reconstruction on the following datasets: cardiac CINE\cite{schlemper2018deep}, coronal knee proton-density (Knee-CPD) and axial knee T2 (Knee-AT2)\footnote{For knee images we used retrospectively down-sampled ground truth images to simulate single-coil reconstruction. 4$\times$ Gaussian variable density Cartesian down-sampling is used for all datasets.}\cite{hammernik2018learning}. 

To show that phase-synthesised MS-COCO provides richer and more variable patch samples for training, we extracted 20k 7$\times$7 patches from each of MRI datasets separately, and extracted 200k patches from 4k MS-COCO images. We calculated average cross-domain patch-wise Euclidean nearest neighbour (NN) distances.

\section{evaluations and conclusions}
\label{sec: evl_conc}
\ninept

\setlength{\tabcolsep}{3pt} 
\def\arraystretch{-1}
\begin{table}[!h]
\label{tbl: recon_value}

\centering
\scalebox{0.83}{
\begin{tabular}{ccccc}

\ninept
\textbf{Train} & \multicolumn{4}{c}{ \textbf{Test} } \\
\hline
& & Cardiac & Knee-CPD & Knee-AT2 \\
\cline{3-5}
\multirow{2}{*}{ Cardiac } & \textit{PSNR} & \colorbox{gray!30}{\textbf{29.95$\pm$1.96}} & 31.52$\pm$3.14 & 33.64$\pm$1.68 \\
& \textit{SSIM} & \colorbox{gray!30}{\textbf{0.95$\pm$0.01}} & 0.95$\pm$0.05 & 0.94$\pm$0.02 \\
\cline{3-5}
\multirow{2}{*}{ Knee-CPD } & \textit{PSNR} & 26.91$\pm$2.47 & \colorbox{gray!30}{33.42$\pm$2.79} & 34.01$\pm$1.66 \\
& \textit{SSIM} & 0.90$\pm$0.03 & \colorbox{gray!30}{\textbf{0.96$\pm$0.05}} & 0.95$\pm$0.01 \\
\cline{3-5}
\multirow{2}{*}{ Knee-AT2 } & \textit{PSNR} & 24.96$\pm$2.56 & 31.78$\pm$3.86 & \colorbox{gray!30}{\textbf{35.34$\pm$1.79}} \\
& \textit{SSIM} & 0.85$\pm$0.05 & 0.95$\pm$0.05 & \colorbox{gray!30}{\textbf{0.96$\pm$0.01}} \\
\cline{3-5}
\multirow{2}{*}{ MS-COCO } & \textit{PSNR} & 29.46$\pm$2.19 & \textbf{33.79$\pm$2.72} & 34.70$\pm$1.74 \\
& \textit{SSIM} & 0.94$\pm$0.01 & \textbf{0.96$\pm$0.05} & 0.95$\pm$0.02 
\end{tabular}}
\caption{Quantitative evaluations of cross-domain reconstruction}
\end{table}

As shown in Table 1, Training with phase-synthesised MS-COCO yields the best overall cross-domain performances. This behaviour is consistent with observations on magnitude-only images\cite{dar2017transfer}. Domain-correct reconstructions are \colorbox{gray!30}{highlighted} for reference.

Table 2 shows that MS-COCO provides the smallest average cross-domain patch-wise NN distances ($p$-value $< 10^{-10}$). This implies that the generalisation ability of our method might be related to large intersections in patch subspaces between domains.

\begin{figure}[!h]
\label{fig: error}
\centering
\scalebox{0.85}{
\begin{minipage}[b]{1\linewidth}
  \centering
  \centerline{\includegraphics[width=8.5cm]{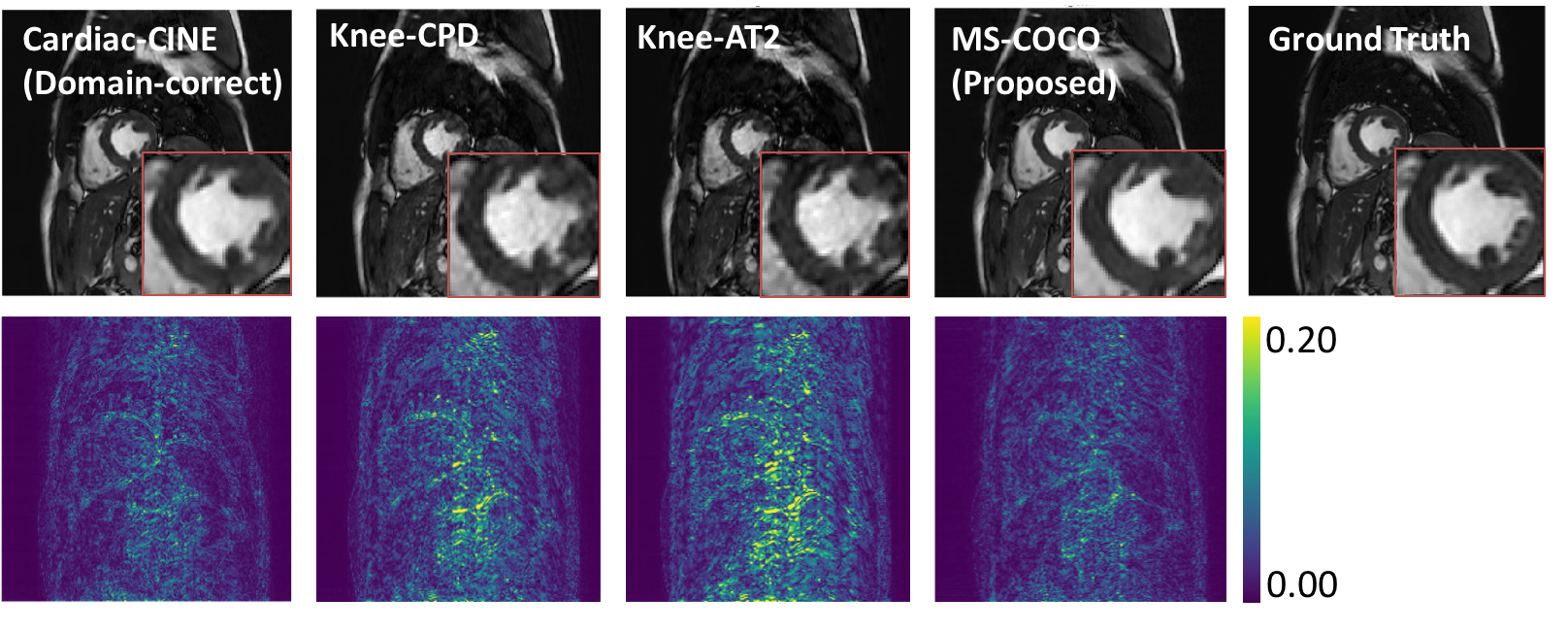}}
  \vspace{0cm}
\end{minipage}
}
\caption{Images for cross-domain reconstruction}
\label{fig:res}
\end{figure}

\setlength{\tabcolsep}{3pt} 
\renewcommand{\arraystretch}{0} 
\begin{table}[!h]
\label{tbl: distance}
    \centering
        \scalebox{0.84}{
        \begin{tabular}{ccccc}
        \textbf{Target} & \multicolumn{4}{c}{ \textbf{Source} } \\
        \hline
        & Cardiac & Knee-CPD & Knee-AT2 & MS-COCO \\
        \cline{2-5}
        Cardiac & - & 0.41$\pm$0.33 & 0.64$\pm$0.54 & \textbf{0.37$\pm$0.28} \\
        Knee-CPD & 2.52$\pm$3.64 & - & 1.38$\pm$1.20 & \textbf{0.84$\pm$0.96} \\
        Knee-AT2 & 2.45$\pm$4.60 & 1.40$\pm$2.32 & - & \textbf{1.35$\pm$2.25} \\
        \end{tabular}
}
    \caption{Average cross-domain patch-wise NN distances}
\end{table}

\bibliographystyle{IEEEbib}
\ninept
\bibliography{refs}
\ninept
\end{document}